\pdfoutput=1
\newcommand\anonymize[1]{[ANONYMIZED]}
\documentclass[11pt]{article}
\usepackage{listings}
\usepackage[final]{acl}

\usepackage{times}
\usepackage{latexsym}
\usepackage{enumitem}
\usepackage[T1]{fontenc}

\usepackage[utf8]{inputenc}

\usepackage{microtype}

\usepackage{inconsolata}

\usepackage{graphicx}
\usepackage{longtable}
\usepackage{supertabular,booktabs}

\usepackage{tipa,tabularx}

\title{Analysis of LLM as a grammatical feature tagger for African American English}

\author{
 \textbf{Rahul Porwal\textsuperscript{1}},
 \textbf{Alice Rozet\textsuperscript{1}},
 \textbf{Jotsna Gowda\textsuperscript{1}},
 \textbf{Pryce Houck\textsuperscript{1}},
 \textbf{Sarah Moeller\textsuperscript{1}},
 \textbf{Kevin Tang\textsuperscript{2,1}}
\\
\\
 \textsuperscript{1}University of Florida, Department of Linguistics, College of Liberal Arts and Sciences,\\\texttt{\{rahulporwal, arozet, jotsna.gowda,phouck,smoeller\}@ufl.edu}\\ 
 \textsuperscript{2}Heinrich Heine University Düsseldorf,\\ Department of English Language and Linguistics, Faculty of Arts and Humanities\\\texttt{kevin.tang@hhu.de}
}

\begin{document}
\maketitle
\begin{abstract}
African American English (AAE) presents unique challenges in natural language processing (NLP) %
This research systematically compares the performance of available NLP models—rule-based, transformer-based, and large language models (LLMs)—capable of identifying key grammatical features of AAE, namely Habitual Be and Multiple Negation. These features were selected for their distinct grammatical complexity and frequency of occurrence. The evaluation involved sentence-level binary classification tasks, using both zero-shot and few-shot strategies. The analysis reveals that while LLMs show promise compared to the baseline, they are influenced by biases such as recency and unrelated features in the text such as formality. %
This study highlights the necessity for improved model training and architectural adjustments to better accommodate AAE's unique linguistic characteristics.
Data and code are available.

\end{abstract}

\section{Introduction}

African American English (AAE) is a low-resource language, facing the challenge of inadequate training data for natural language processing (NLP) \citep{blodgett-etal-2018-twitter,LucaStreiter2003}. While efforts have shown promise in improving NLP performance on AAE \citep{dacon2022deep,masis2023investigating}, progress is limited. Consequently, AAE lacks access to the same host of language-specific tools available to varieties such as Mainstream American English (MAE). With the increasing use of large language models (LLM) to perform annotation tasks, studies found that LLMs performs well for high-resource languages such as English but the performance drops for non-English languages  \citep{jadhav2024limitationsllmannotatorlow,pavlovic-poesio-2024-effectiveness}. This raises the question of whether LLMs would be able to handle AAE, a variety of English that is low-resource.

This paper evaluates the ability of different NLP systems to recognize distinctive AAE grammatical features, comparing a rule-based model, a transformer-based model, and LLM on the same tasks, specifically tagging distinctive AAE grammatical features. It compares LLM zero- or few-shot classification. Data and code are available.\footnote{\url{https://github.com/tang-kevin/AfricanAmericanEnglishTagger}}

The systems are evaluated on two distinctive AAE features as a test case: Habitual Be and Multiple Negation. Habitual Be is relatively rare and difficult to model while Multiple Negation is prevalent in AAE and occurs across various non-standard English dialects. %
These features provide different frequency of occurrence, extremity in terms of grammatical complexity, and possible prior exposure by LLMs. 
Furthermore, Habitual Be and Multiple Negation is found in 27\% and 81\% of English varieties, respectively.\footnote{\href{https://ewave-atlas.org/}{https://ewave-atlas.org/}} 

This work poses these questions:

\begin{itemize}[noitemsep,topsep=0pt]
    \item How LLMs compare to rule-based models or trained-from-scratch Transformer models at identifying AAE grammatical features?
    \item How does recency or order of examples influence LLM performance?
    \item How do factors of transcribed spoken language influence LLM performance at identification of AAE?
\end{itemize}

\section{Related Work}

NLP research for AAE have explored social media use \citep{blodgett-etal-2018-twitter}, POS-tagging \citep{dacon2022deep,JorgensenHS16}, hate speech classification \citep{Harrisetalhatespeechgrammar2022,SapCGCS19}, ASR \citep{koenecke2020racial,Martin_Tang_RacialBias_HabBe_Interspeech2020_2020}, dialectal analysis
\citep{BlodgettGO16,dacon2022deep,Stewart14} and feature detection \citep{masis-etal-2022-corpus,santiago_disambiguation_2022,previlon-etal-2024-leveraging}. Our work highlights the lack of attention to AAE's distinctive grammatical structure that can be used to identify its usage accurately. Methods mitigating bias in NLP often neglect AAE's grammatical features in favor of lexical choice 
\citep{BarikeriLVG20,ChengK022,GarimellaMA22,hwang-etal-2020-towards, KiritchenkoM18,maronikolakis-etal-2022-analyzing,silva-etal-2021-towards}. Some research removes AAE's morphological features \citep{tan-etal-2020-mind} or translates between MAE and AAE \citep{ziems-etal-2023-multi}.

\section{AAE Grammatical Features}

Our work focuses on two morphosyntactic features that characterize AAE and distinguish it from MAE. The features provide extremity in terms of grammatical complexity and possible prior exposure by LLMs. Multiple Negation is prevalent in our AAE dataset and its occurrences are uniform enough to captured with a rule-based model. Habitual Be is comparatively rare. Its complexity requires a combined rule-based and probabilistic baseline from previous work  \cite{santiago_disambiguation_2022,MoellerDBPTang_RiCL_2024,previlon-etal-2024-leveraging}.  %

\subsection{Habitual Be}

Although relatively infrequent, habitual \textit{be} is employed regularly by AAE speakers \citep{BlodgettGO16} as well as speakers of 90 other Englishes \citep{kortmann2020syntactic}. It is an aspectual marker denoting a recurring, or habitual, action \citep{green_african_2002, fasold_tense_1969}. In contrast to other instances of ``be", the habitual \textit{be} is never changes form. As shown in the MAE sentence below, non-habitual ``be'' must agree with ``I" resulting in ``I am" rather than ``I be". Additionally, the adverb "usually" in indicates the event is recurring. Habitual Be, as in the AAE exmaple, does not require the adverb to specify habituality.

\begin{enumerate}
    \item[]\label{ex:habbe} \textbf{AAE}: I be in my office by 7:30.\\
        \textbf{MAE}: I am usually in my office by 7:30.
\end{enumerate}
\subsection{Multiple negation}

Multiple Negation, or Double Negative or Negative Concord, is characterized by multiple negative words such that each negator confirms an overall negative meaning of a single linguistic constituent as in the AAE example below. This contrasts to ``I didn't ask you not to come'' where each negative word belongs to, and negates, a different clausal constituent. Multiple Negation %
was used in Old English \citep{Kallel2011TheLO} and is today a recognized feature of several English varieties but is absent from MAE \citep{martinez2003}, as shown below. 

\begin{enumerate}
    \item[]\label{ex:multneg} \textbf{AAE}: I ain't step on no dog.\\
         		\textbf{MAE}: `I didn’t step on a dog.'
\end{enumerate}

\section{Data \& Corpus} %

All data used for the training, testing, and analysis of the included models was transcribed, tagged, and annotated for AAE features by hand from texts in the Joel Buchanan Archive,\footnote{Part of the Samuel Proctor Oral History Program (SPOHP) at the University of Florida \url{https://oral.history.ufl.edu/projects/joel-buchanan-archive/}} a collection of oral histories gathered primarily from African Americans, or in The Corpus of Regional African American Language (CORAAL) \citep{farrington_corpus_2021}. 
Each example is sourced from an interview with an AAE speaker and contains a key component of the grammatical feature, i.e. `be' for habitual be and at least one instance of a negative word (i.e. not, never, etc.) for multiple negation. For simplicity sake, Non/Habitual Be examples were limited to sentences with a single ``be''. 

Sentences in the the databases are marked by humans for the presence of AAE grammatical or phonological features. For example, if a sentence contains a habitual ``be'', its line is tagged with a `1' in a column corresponding to the feature, otherwise `0'. %
All annotators had to qualify after training in recognition of AAE linguistic features with a high level of accuracy. For detailed annotation guidelines and training materials, see \citet{moeller_tang_ALFAH}. Each transcribed interview is annotated by at least one annotator, and through the tagger and model creation process, the annotations of sentences used for data input or analysis are reviewed for accuracy. The labels assigned by human annotators are treated as gold standard when analyzing the performance of the models.  %

Training/testing sets are generated for individual grammatical features. %
Using sentences from annotated interviews, datasets are created with a predetermined ratio of `positive examples' %
that contain the grammatical feature, to negative examples, which do not. %
This guarantees that the models we build do not experience an increase in perceived accuracy from their ability to word search but rather can truly distinguish AAE from MAE.  %

\section{Model architecture and configurations} 

We compared two LLMs, one Transformer model, and one rule-based model. The task was set up as binary classification on sentences such that each model tags a given sentence as either positive or negative for a specific AAE feature of interest. For the sentence-level binary classification tasks, each task involved analyzing, training, or prompting these models on batches of sentences, which were processed with a consistent prompt format in the LLMs. 

\subsection{Large Language Models}

The language models used in this study were chosen based on their suitability for sentence-level binary classification tasks, given the context and parameter constraints. Each model was configured with a consistent set of hyperparameters and prompt structures to enable a fair evaluation of their performance across various sentence structures. 

\paragraph{OpenAI -- GPT.}
OpenAI’s \texttt{gpt-4o-mini} model, a compact language model containing approximately 8 billion parameters, was accessed through the OpenAI REST API with a configuration set to a temperature of 0.7 and a top-p value of 0.9, providing a balance between response coherence and variability. 

\paragraph{Meta -- LLaMA}
Meta’s \texttt{LLaMA 3-8B-Instruct} has a similar architecture and parameter size (8 billion parameters) to OpenAI’s \texttt{gpt-4o-mini}. The \texttt{LLaMA} model was accessed through the \texttt{Huggingface} \texttt{transformers} library %
executed using \texttt{PyTorch} v12.4 with NVIDIA CUDA support. The experimental settings, including temperature (0.7) and top-p (0.9), were set identical to those used for \texttt{gpt-4o-mini} to maintain consistency in response characteristics.

\subsection{Baselines} %

We compared the LLMs models previously built to identify specific AAE grammatical features. 
The Habitual Be baseline is a Transformer model based on the work in \citet{previlon-etal-2024-leveraging}. The inputs to the model include n-grams windows around ``be'' along with predictions of the feature's presence based on part of speech (POS) windows and contextual syntactic structures. %
Since training data was limited, we averaged results across a 10-fold cross validation and augmented the dataset. Approximately 3,500 more Habitual Be sentences with the method described in \citet{santiago_disambiguation_2022} to provide an evenly balanced the dataset. These augmented sentences structurally mirror the Habitual Be sentences in the interview transcripts %
The data is split into 10 folds, each containing 774 samples (401 non-habitual / 373 habitual).

\setlength{\tabcolsep}{2pt}
\begin{table}
\centering
\begin{tabular}{|l|c||c|}
\hline
   & \textbf{Multiple Negation} & \textbf{Habitual Be}  \\  
  \hline                
  \textbf{$+$} & 4,009 & 37 \\
  \textbf{$-$} & 3,730 & 161 \\
  \hline                
  \textbf{Total} & 7,739 & 198 \\
  \hline 
\end{tabular}
\caption{The datasets. In the table, the `+' class indicates presence of the feature and `--' indicates the examples contain a key component (e.g. ``be'') but do not exhibit the grammatical feature.}
\label{tab:data}
\end{table}

For the multiple negation tagger, we analyzed the patterns found in the annotated data and wrote a rule-based model that predicts the presence or absence of multiple negation in a sentence containing at least one negative word. %
The tagger identifies negators (e.g. ``not'', ``nothing'') in a sentence and searches for another negator within the clause boundaries. Clause boundaries are defined by punctuation or conjunctions. If a second instance is found, the model tags it postive for Multiple Negation. The dataset size is described in Table \ref{tab:data}. %

\section{Prompt Design}

In LLMs such as GPT4o, prompt design plays a crucial role. %
Prompt length and structure can impact the responses generated, making it important to optimize prompt formulation for specific tasks. We analyzed variations in prompt instructions. %

The same input phrased differently can result in outputs of varying lengths and formats. For instance, a prompt like:

\begin{lstlisting}
"Classify the sentence '{sentence}' as 'habitual be' or 'non-habitual be' in one word:"
\end{lstlisting}
results in the desired concise response which contains only the classification label. Without specifying the "in one word" constraint, a prompt like:

\begin{lstlisting}
"Classify the sentence '{sentence}' as 'habitual be' or 'non-habitual be':"
\end{lstlisting}
often leads to the model adding unstructured text such as \texttt{"This sentence uses 'be' in a habitual sense because it describes a repeated action"}. %
This behavior arises because the absence of strict word count or formatting instructions leaves the model room to interpret the instructions, which defaults to more verbose responses. %
The model may include justifications unless the prompt is explicitly constrained. Other prompts like:

\begin{lstlisting}
"Determine if '{sentence}' is 'habitual be' or 'non-habitual be'."}
\end{lstlisting}
might yield an answer like, \texttt{"This is a habitual use because the context suggests a repeated action,"} instead of just the classification term. %

The simpler and more precise the instructions, the easier it is for the model to generate structured and predictable outputs, thereby reducing the need for post-processing. Prompt engineering reduces the need for complex post-processing, making the model more effective for real-world applications.

The prompt generation function transitioned to constructing a set of classification tasks for the model. In our experiments, the prompt for Zero Shot followed a template of:

\begin{lstlisting}
"[batch index]. Classify the sentence '[Sample Sentence]' as '[Feature A]' or '[Feature B]' in one word while preserving the numbering at the start of the prompt."
\end{lstlisting}

The prompt for Few Shot followed these steps:
\begin{enumerate}
    \item Initiate with a predefined text: ``I have given a few classified train examples,'' setting the context for the model.
    \item Append selected training example to the prompt, formatted as: 
    \begin{lstlisting}
    "Sentence: [sentence from data]
    Label: [corresponding label]"
    \end{lstlisting}
    Each example is clearly marked with its gold label, e.g. `habitual be` or `non-habitual be`. %
\end{enumerate}

\section{LLM Compared to Baselines}

To answer our first question regarding the performance of LLMs on AAE features compared to older models, we compare each baseline model to the two LLMs, using the classification report from \texttt{sklearn} \citep{pedregosa2011scikit}. We chose $F_1$-score weighted by class size as our primary metric for evaluation as opposed to accuracy due to the imbalanced class sizes. %

We found that zero and few shot LLM models performed worse compared to the baselines on both task. It underperformed the Habitual Be baseline recall by about -0.02  and overall $F_1$-score by -0.18. 
Compared to our multiple negation baseline tagger, the LLM model performed worse. It underperformed multiple negation recall by about -0.08 and overall $F_1$-score by -0.16. 
In order to understand the source of error within the LLM models, we tested two  hypotheses -- recency bias and formality bias, described in the section \ref{sec:llmbiase}.

\subsection{Results of Baselines} 

These are the results against which we evaluated the LLM models.

\paragraph{Results of Habitual Be Baseline}
The Habitual Be baseline model was evaluated as an average of the overall $F_1$-score of the 10 folds.
The model returned, on average, a 0.88 recall and a 0.92 overall $F_1$-score on the habitual class. %

\paragraph{Results of Multiple Negation Baseline}
The Multiple Negation baseline model achieved a recall of 1.00 and a 0.99 overall $F_1$-score. This means that all sentences containing multiple negation were correctly classified. Only one non-feature sentence was tagged incorrectly.

\subsection{Results of Large Language Models}

{
\setlength{\tabcolsep}{2pt}
\begin{table}
\begin{tabular}{|l|cl|cl||cl|cl|}
\hline
  & \multicolumn{4}{c||}{\textbf{Multiple Negation}}                                          & \multicolumn{4}{c|}{\textbf{Habitual Be}}  \\  \hline                

                & \multicolumn{2}{c|}{GPT}                              & \multicolumn{2}{c||}{\texttt{LLaMA}}       & \multicolumn{2}{c|}{GPT}                                & \multicolumn{2}{c|}{\texttt{LLaMA}}       \\ 
\hline
\textbf{} & \multicolumn{1}{c|}{Zero} & \multicolumn{1}{c|}{Few}  & \multicolumn{1}{c|}{Zero} & Few  & \multicolumn{1}{c|}{Zero}  & \multicolumn{1}{c|}{Few}   & \multicolumn{1}{c|}{Zero} & Few  \\ \hline
$P_+$       & \multicolumn{1}{c|}{0.78} & \multicolumn{1}{l|}{0.47} & \multicolumn{1}{c|}{0.94} & \textbf{1.00} & \multicolumn{1}{c|}{0.67} & \multicolumn{1}{l|}{\textbf{0.69}} & \multicolumn{1}{c|}{0.21} & 0.48 \\ \hline
$P_-$       & \multicolumn{1}{c|}{\textbf{1.00}} & \multicolumn{1}{l|}{0.98} & \multicolumn{1}{c|}{0.38} & 0.56 & \multicolumn{1}{c|}{\textbf{0.68}} & \multicolumn{1}{l|}{0.83} & \multicolumn{1}{c|}{0.55} & 0.66 \\ \hline
$R_+$        & \multicolumn{1}{c|}{\textbf{1.00}} & \multicolumn{1}{l|}{0.94} & \multicolumn{1}{c|}{0.72} & 0.35 & \multicolumn{1}{c|}{0.65} & \multicolumn{1}{l|}{\textbf{0.86}} & \multicolumn{1}{c|}{0.04} & 0.73 \\ \hline
$R_-$        & \multicolumn{1}{c|}{0.94} & \multicolumn{1}{l|}{0.77} & \multicolumn{1}{c|}{0.78} & \textbf{1.00} & \multicolumn{1}{c|}{0.70} & \multicolumn{1}{l|}{0.63} & \multicolumn{1}{c|}{\textbf{0.88}} & 0.39 \\ \hline
$F_1w$        & \multicolumn{1}{c|}{\textbf{0.95}} & \multicolumn{1}{l|}{0.82} & \multicolumn{1}{c|}{0.76} & 0.59 & \multicolumn{1}{c|}{0.67} & \multicolumn{1}{l|}{\textbf{0.74}} & \multicolumn{1}{c|}{0.41} & 0.53 \\ \hline
\end{tabular}
\caption{Classification performance of Multiple Negation and Habitual Be detection using the GPT model and the \texttt{LLaMA} model using zero-shot and few-shot prompting strategies. $P_+$ and $P_-$: precision of the positive class (feature present) and of the negative class (feature absent); $R_+$ and $R_-$: recall of the positive and negative classes; $F_1w$: weighted $F_1$}
\label{tab:zerovsfew}
\end{table}
}

The LLM models results with zero shot and few shot prompting strategies %
are shown in Table \ref{tab:zerovsfew}. In the table, the `+' class indicates presence of the feature and `--' indicates absence of the feature. 

Overall, the GPT model performed better than \texttt{LLaMA} both in terms of memory constraints and rate limit as well as quality of detection, compared to the \texttt{LLaMA} model. It seems evident that the \texttt{LLaMA} model has not been trained on a lot of data from African American English. Showing it a few examples helped the performance immensely. This shows that fine tuning the model with more data from African American English could improve its performance. The performance of the GPT model was quite contrasting for the Habitual Be and Multiple Negation feature detection. %
This can be attributed to the syntactic structural complexity of the two features. Multiple Negation is easier to detect by simple count of lexical items and hence, LLMs do not need any kind of examples to be able to detect it. Habitual Be, on the other hand, is difficult to detect without some representation of the underlying grammatical structure. %
Without having the context or extra data regarding this, LLMs find it harder to detect it. Overall, for both the features, LLMs seem to be biased towards predicting the presence of a feature.

\subsubsection{Habitual Be}
\label{subsubsection:habbe}

For Habitual Be, the GPT model performance for zero-shot prompting was quite poor compared to few shot, particularly when we compare between both prompting styles the numbers for the feature (habitual) class recall, that is, how many true Habitual Be samples the LLM was able to capture. This can be attributed to the complex and nuanced nature of Habitual Be. The increase in performance of the GPT model from zero to few shot prompting can also be primarily attributed to the increase in recall of the habitual class. We speculate that when the LLM is exposed to a few examples, it develops a better understanding of what syntactic structure to look for in order to detect Habitual Be. In the case of the GPT model, the precision stays almost the same. Meanwhile, the recall of the non-habitual class decreases. It is possible that with prompting, the model may be more biased towards detecting that a sentence contains a habitual be feature, leading to both higher habitual recall and lower non-habitual recall. 

The performance with the \texttt{LLaMA} model could not be computed on all the 7,739 samples due to memory constraints and rate limits. So we performed the experiment with a reduced total dataset of 1,000 samples (430 habitual and 570 non-Habitual). We utilized a similar 10 fold data split as the GPT model but because the number of samples for each fold was lower, we decided to generate a combined classification report for all the 1000 samples. The \texttt{LLaMA} model is unable to detects Habitual Be poorly unless being nudged by a few examples. 

\subsubsection{Multiple Negation}

For the multiple negation model, we tested the performance of the GPT model on a dataset containing 198 samples, out of which 162 contain Multiple Negation and the remaining 36 do not. %
On GPT, the zero-shot prompting produced better results compared to few shot. This performance difference is prominent when we look at the dip in precision scores for both the feature (multiple negation) and non-feature (not multiple negation) class. The high recall and low precision of the LLM is likely because the LLMs are biased towards predicting a  sentence has multiple negation if there are two or more negative words in a sentence. Multiple Negation is a simpler feature to detect, requiring only that there are two negators within a single clause. %
It is possible, albeit with less accuracy, to identify Multiple Negation with high likelihood based solely on the number of negative words the LLM sees in a sentence. This may be one reason the model tends to have a higher recall and lower precision. This nuance whether any two negative words in the same sentence are not also in the same clause, can easily be overlooked, leading to sentences being mistakenly classified as having the feature. The precision score falls further with few shot prompting, likely due to the model overfitting to the few extra samples that it learns in the prompt. For a simple-to-detect feature with nuanced underlying grammatical structure, like multiple negation, it may be counterproductive to feed the LLM more examples. 

We also tested the performance of \texttt{LLaMA} model for multiple negation with the same dataset of 198 samples. The \texttt{LLaMA} model performed poorly compared to GPT overall, but its precision for the feature class is much higher than the GPT model. This appears to due more to the model's bias towards predicting that the sentences do not contain multiple negation. The lower recall for \texttt{LLaMA} can be attributed to the much smaller training dataset compared to GPT. We notice a similar trend with \texttt{LLaMA} for Multiple Negation that we saw with GPT--that performance with zero shot is better than few shot.

\section{Potential LLM Biases}
\label{sec:llmbiase}

To understand the cause for the bias towards predicting the presence of an AAE feature and to answer our last three questions, we analyzed the LLM predictions and saw three major factors influencing them. %

\textbf{Recency bias} - Recency bias occurs when the LLM's predictions for the current sentence can be affected by the previous sentence's prediction. This was inspired by recent work on testing human cognitive effects in LLMs \citep{Shaki_2023,clark-etal-2025-linear}.
As the input to the LLMs in our experiment were in batches, recency bias from the previous sentences and predictions in the same batch can influence the model's predictions.\footnote{Other studies of LLMs have used the term recency bias to describe when models prefer information processed more recently in their input data \citep{guo2024serialpositioneffectslarge} and to describe when long context language models would prefer information presented at the end of training documents \citep{Peysakhovich2023AttentionSC}.
While this use of the term is similar to the one used here, the difference lies in how ours refers to the influence from the previous sentences and their predictions in the same batch.} We observed this effect where the model incorrectly predicted that a sentence contained multiple negation.

\textbf{Formality bias} - Under informality we group run-on, incomplete, fragmented, and disjointed sentences that may have either excessive or unusual punctuation. Some of these issues stem from working with transcribed oral speech rather than written English but these issues may have implications for working with non-standard writing common on social media. We observed that for some longer sentences and disjoint sentences where there is a higher number of punctuations like semi-colons and commas, the LLM performance degraded. Even in the case of sentences containing grammatical or orthographic errors, the model also tends to perform poorly.  %

\textbf{AAE bias} - We observed many examples where the model predicted Multiple Negation or Habitual Be when there was presence of a different AAE grammatical feature. But this bias also arises when any non-standard feature is present, whether or not it is characteristic of AAE, for example, the presence of the word ``ain't". This seems to be a major reason for LLMS incorrectly identifying AAE structure. LLMs predicted the sentences in the examples below as containing multiple negation, but they do not. The first sentence does contain another AAE grammatical feature Person Number disagreement (``We was" rather than ``we were"). However, in the second example, ``ain't", although commonly used in AAE, is also used regularly by speakers of many English varieties, including speakers of MAE. Based on our analysis, it seems that the presence of any feature not considered standard in written English biases the model to predict the presence of an AAE grammatical structure.  %
\begin{enumerate}
    \item[]\label{ex:aaefeaturebias1} \textbf{Example 1}: We was in Pentecost holiness and I wasn't allowed to smoke.
    \item[]\label{ex:aaefeaturebias2} \textbf{Example 2}: Because he ain't been back to finish yet.
\end{enumerate}

Below we systematically examine the Recency bias and Formality bias to determine whether these factors do seem to influence LLMs and to quantify their impact on the task of detecting multiple negation and habitual be features. We felt there was insufficient evidence to properly characterize the issues we categorize as AAE bias.

\subsection{Hypothesis 1: Recency Bias}\label{subsection:hypo1} %
\subsubsection{Methodology}\label{subsubsection:hypo1method}

Recency bias in the LLMs' predictions was assessed using logistic regression fitted using Maximum Likelihood Estimation on aforementioned experiments, where the relevant input sentences were presented to the LLM in a random order without probing for any specific error. The regression was conducted using the Python packages `statsmodels', `numpy', and `pandas'. 
The dependent variable is the model prediction (presence or absent of a feature). The independent variable of interest is the presence of a recency bias. It is operationalised as the proportion of the predicted values of the last five samples ($N_{-5}$,...$N_{-1}$) that match the predicted value of the given sample ($N$). The recency bias can either positively or negatively affect the prediction. A positive effect indicates the model prefers using recently predicted values, while a negative effect indicates the model avoids using recently predicted values. Ground truths were treated as an additional independent variable to control for the expected performance of the model, since the predicted values are expected to positively correlate with the ground truth.

{ 
\setlength{\tabcolsep}{1.5pt}
\begin{table}[h!]
    \centering
\begin{tabular}{|c|cc||cc|}
\hline
 & \multicolumn{2}{c||}{\textbf{Multiple Negation}}                    & \multicolumn{2}{c|}{\textbf{Habitual Be}}                          \\ \hline
                        & \multicolumn{1}{c|}{GPT}              & \multicolumn{1}{c||}{\texttt{LLaMA}} & \multicolumn{1}{c|}{GPT}              & \multicolumn{1}{c|}{\texttt{LLaMA}} \\ \hline
Recency $\hat{\beta}$
                               & \multicolumn{1}{c|}{-8.34}          & -4.87                    & \multicolumn{1}{c|}{-0.21}          & -16.52                   \\ \hline
Recency \textit{p}                           & \multicolumn{1}{c|}{\textless{}0.001}          & \textless{}0.001                    & \multicolumn{1}{c|}{0.42}          & \textless{}0.001                   \\ \hline

Ground $\hat{\beta}$                             & \multicolumn{1}{c|}{5.59}          & 2.17                   & \multicolumn{1}{c|}{1.70}          & 1.48                   \\ \hline
Ground \textit{p}                                  & \multicolumn{1}{c|}{\textless{}0.001}          & \textless{}0.001                    & \multicolumn{1}{c|}{\textless{}0.001}          & 0.15                   \\ \hline  \hline

Pseudo R\textsuperscript{2} & \multicolumn{1}{c|}{0.77}             & 0.28                       & \multicolumn{1}{c|}{0.12}             & 0.90                       \\ \hline
\end{tabular}
\caption{Recency bias -- Logistics regression summaries of Multiple Negation and Habitual Be detection using GPT and \texttt{LLaMA} models with zero-shot prompting. Recency/Ground $\hat{\beta}$ and Recency/Ground \textit{p} denote the coefficient and the p-value of the recency bias variable and the ground truth variable.}
\label{tab:formality-logsummaries}
\end{table}
}

\subsubsection{Analysis: Zero-shot Prompting}
The logistic regression analysis reveals whether the recency bias has an effect on sentence classification of two AAE features under each prompting paradigm. The regression summaries can be found in Table \ref{tab:formality-logsummaries}.

\textbf{Habitual Be detection with GPT}
The regression analysis reveal that the $\hat{\beta}$ for recency bias to be -0.21 and it is not statistically significant (\textit{p}-value = 0.42). This indicates that, in the zero-shot setting, recency bias does not have a meaningful impact on the model's ability to classify Habitual Be. Unsurprisingly, the control variable (ground truth) has a positive effect on ($\hat{\beta}$ = 1.740, \textit{p}-value \textless{}0.001). %

\textbf{Multiple Negation detection with GPT}
For the zero-shot scenario for Multiple Negation feature in the GPT Model, the results indicate a strong \textit{negative} relationship between recency bias and model performance ($\hat{\beta}$ = -8.34, \textit{p}-value \textless{}0.001). This suggests that the model avoids generating a predicted value that it recently used. Unsurprisingly, the control variable (ground truths) has a positive effect on the model's overall fit ($\hat{\beta}$ = 5.59, \textit{p}-value \textless{}0.001). %

\textbf{Multiple Negation \& Habitual Be detection with \texttt{LLaMA}}
Recency bias was found also with both Multiple Negation and Habitual Be detection using \texttt{LLaMA}.
Similar to the GPT model, there is a strong \textit{negative} relationship between recency bias and the \texttt{LLaMA} model's performance to detect multiple negation ($\hat{\beta}$ = -4.88, \textit{p}-value \textless{}0.001). Whereas the GPT model does not have a recency bias when detecting Habitual Be, the \texttt{LLaMA} model has such a bias ($\hat{\beta}$ = -16.52, \textit{p}-value \textless{}0.001).

With the exception of the GPT model, the recency bias is stronger with Habitual Be than with multiple negation. The exact nature of this difference might be due to the fact that Habitual Be has a range of complex syntactic environments \citep{previlon-etal-2024-leveraging}, while Multiple Negation, although a syntactic feature, is easier to describe in terms of a limited list of words. %
\texttt{LLaMA} would typically mark any sentence that contained more than one negator word as being Multiple Negation, with this often being a correct assumption. On the other hand, the Habitual Be relies on more experience interpreting non-standard American English to recognize. Given that the zero-shot experiments do not provide examples, one can speculate that \texttt{LLaMA} is using previous inputs and their predictions to guide future prediction due to a lack of sufficient training on AAE \citep{martin2022phd}.

\subsubsection{Additional analysis: Few-Shot prompting}

\begin{table}[h]
\centering
\begin{tabular}{|c|cc||cc|}
\hline
 & \multicolumn{2}{c||}{\textbf{Multiple Negation}}   & \multicolumn{2}{c|}{\textbf{Habitual Be}}  \\
 \hline
   & \multicolumn{1}{c|}{GPT}              & \multicolumn{1}{c||}{\texttt{LLaMA}} & \multicolumn{1}{c|}{GPT}              & \multicolumn{1}{c|}{\texttt{LLaMA}} \\
\hline
Recency $\beta$ & \multicolumn{1}{c|}{-12.36} & 1.22 & \multicolumn{1}{c|}{-1.03} & 1.81 \\
\hline
Recency $p$ & \multicolumn{1}{c|}{<0.001} & 0.112 & \multicolumn{1}{c|}{0.003} & <0.001 \\
\hline
Ground $\beta$ & \multicolumn{1}{c|}{4.13} & 30.81 & \multicolumn{1}{c|}{3.42} & 0.93 \\
\hline
Ground $p$ & \multicolumn{1}{c|}{<0.001} & 1.000 & \multicolumn{1}{c|}{<0.001} & 0.318 \\
\hline
\hline
Pseudo $R^2$ & \multicolumn{1}{c|}{0.59} & 0.21 & \multicolumn{1}{c|}{0.33} & 0.10 \\ \hline
\end{tabular}
\caption{Few-Shot Prompting: Logistics regression summaries of Multiple Negation and Habitual be using GPT and LLaMA. Re-
cency/Ground $\hat{\beta}$ and Recency/Ground p denote the co-
efficient and the p-value of the recency bias variable and
the ground truth variable.}
\label{tab:regression_results}
\end{table}

\textbf{Habitual Be detection with GPT} 
In contrast, the few-shot prompting results indicate a significant negative impact of recency bias on prediction accuracy ($\hat{\beta}$ = -1.03, \textit{p}-value = 0.003). This substantial effect suggests that the few-shot model is considerably influenced by the timing and sequence of input data, potentially leading to biased predictions based on recent but not necessarily representative samples. The true label in this scenario also exhibits a strong positive influence ($\hat{\beta}$ = 3.42, \textit{p}-value < 0.001), reinforcing the label's dominant role in driving predictions. With a higher Pseudo R² of 0.33, the few-shot model demonstrates a better fit than the zero-shot model. %

\textbf{Multiple Negation detection with GPT} 
In the few-shot prompting scenario for the Multiple Negation feature, the analysis presents an even more pronounced negative impact of recency bias ($\hat{\beta}$ = -12.35). This stronger negative effect could be attributed to the few-shot model's reliance on a limited number of examples, which may accentuate the influence of recent inputs, thereby skewing prediction outcomes significantly. The presence of a lower, yet still positive, coefficient for the ground truth variable ($\hat{\beta}$ = 4.13) underscores a consistent trend where the intrinsic attributes of labels influence model predictions, but the effect of recency bias remains a dominant factor, adversely affecting prediction accuracy.

\paragraph{Few-Shot Prompting with \texttt{LLaMA}}
Some additional testing examined the relationship between \texttt{LLaMA}'s strong recency bias and the examples provided in few-shot testing more closely. A set of ten prompts were chosen that the LLM failed on previous experiments, five with true labels `0' and five with true labels `1'. First, these were presented in alternating order, then with all consecutive 1-labeled sentences followed by all 0-labeled sentences, and finally with this ordering reversed. These two experiments were run again with the batch size increased to 30. Each of these four tests were run six separate times with the same set of examples ordered differently to measure if the order of provided samples impacted recency bias.
Overall, few-shot prompting demonstrated more sensitivity to recency in provided examples than sentence history, while still being much more accurate than zero-shot counterparts even when negatively impacted. The ordering of example sentences had a substantial impact on prediction accuracy while the ordering of input sentences largely did not, further suggesting that text within the most recently processed five to ten sentences have a strong impact on \texttt{LLaMA}'s predictions when the criteria for a given feature's presence is not clear.

\subsection{Hypothesis 2: Formality bias} 

Our second hypothesis is that the LLM models are hyper-sensitive to deviations from formal written texts with the according sentence structure and style. Therefore, they flag excessive or unusual punctuation, missing subjects, or run on sentences as AAE features. 

We define 'excessive or unusual' as any repetitive or unnecessary punctuation that leads to syntactic errors or sentence disjunction. We defined `run-on sentences' as either overly long sentences without standard punctuation or multiple disjointed thoughts connected via punctuation. 

\begin{enumerate}[noitemsep,topsep=0pt]
    \item[]\textbf{Example}: "And so they had really- you know; middle-class- they hadn't encountered any real racism."
    \item[]\textbf{Example}: "And so that made me really feel good because you did not have to do this and I thought it was the greatest gesture that you could have come and share this information with us of how to move forward and I don’t know if Florida would have done that even though— I just don’t if they would have taken the time to do that."
\end{enumerate}

In the first example above, the speaker makes false starts and switches between multiple trains of thought, leading to a disjointed sentence. The sentence is not particularly long, but the phrasing and intention is repeatedly broken up, making it more difficult for a model to process and identify grammatical features. This arises from accurately transcribed spoken language.
In the second example, the sentence is excessively long and run-on, with pronouns referring to entities that are several clauses back. Similar to the first example, it shows people's natural conversation process. This makes the data, which is sourced from interview transcriptions, more difficult to understand and analyze.

\subsubsection{Methodology}
Similarly to Section \ref{subsubsection:hypo1method}, we employed a logistic regression analysis. The dependent variable is the predicted value of the models. To examine the hypothesis, the variable of interest is the presence of a deviation from formal written text. Each sentence was manually tagged as `1' if it contained excessive or unusual punctuation, missing subjects, or run-on sentences, and as `0' if it did not. Ground truths are whether the habitual be or multiple negation features were actually present, were treated as control variables in the regression. Due to the size of the Habitual Be dataset, only fold 2 was tested in this hypothesis. It is a fair representation of the entire Habitual Be dataset because it yielded the median k-fold results from the zero-shot GPT model.

{ 
\setlength{\tabcolsep}{1.5pt}
\begin{table}[h!]
    \centering
\begin{tabular}{|c|cc||cc|}
\hline
 & \multicolumn{2}{c||}{\textbf{Multiple Negation}}                    & \multicolumn{2}{c|}{\textbf{Habitual Be}}                          \\ \hline

                        & \multicolumn{1}{c|}{Zero}         & \multicolumn{1}{c||}{Few} & \multicolumn{1}
                        {c|}{Zero}             & \multicolumn{1}{c|}{Few} \\  \hline

Formality $\hat{\beta}$
        & \multicolumn{1}{c|}{0.17}          & 0.89                   & \multicolumn{1}{c|}{0.10}          & 0.96                  \\ \hline
Formality \textit{p}                           & \multicolumn{1}{c|}{0.78}          & 0.02                    & \multicolumn{1}{c|}{0.53}          &      \textless{}0.001             \\ \hline

Ground $\hat{\beta}$                             & \multicolumn{1}{c|}{5.39}          & 3.67                 & \multicolumn{1}{c|}{1.42}          & 1.77                 \\ \hline
Ground \textit{p}                                  & \multicolumn{1}{c|}{\textless{}0.001}          & \textless{}0.001                    & \multicolumn{1}{c|}{\textless{}0.001}          & \textless{}0.001                  \\ \hline  \hline

Pseudo R\textsuperscript{2} & \multicolumn{1}{c|}{0.54}             & 0.26                     & \multicolumn{1}{c|}{0.09}             & 0.17                      \\ \hline
\end{tabular}
\caption{Formality bias -- Logistics regression summaries of Multiple Negation and Habitual Be detection using GPT with zero-shot and few-shot prompting strategies. Formality/Ground $\hat{\beta}$ and Formality/Ground \textit{p} denote the coefficient and the p-value of the formality bias variable and the ground truth variable.}
\label{tab:recencybias-logsummaries}
\end{table}
}

\subsubsection{Analysis}

The logistic regression summaries can be found in Table \ref{tab:recencybias-logsummaries}. Unsurprisingly, the ground truth variable has a positive and significant effect in all four models. Turning to the bias variable, the GPT model with zero-shot prompting does not suffer from a formality bias with both features.  While the formality bias variable in GPT has a positive $\hat{\beta}$ of 0.17 and 0.10 for Multiple Negation and Habitual Be respectively, it was statistically insignificant in both cases (\textit{p}s > 0.05). On the other hand, the GPT model with few-shot prompting does suffer from a formality bias with both features. Across the features, the bias variable has a significant effect only for the few-shot strategy as opposed to the zero-shot. This suggests that the influence of formality bias depends on the examples given to the model. 

The bias has a stronger effect on Habitual Be ($\hat{\beta}$ = 0.96, \textit{p} \textless{}0.001) than on Multiple Negation ($\hat{\beta}$ = 0.86, \textit{p} \textless{}0.001). This pattern of Habitual Be being more susceptible to a bias was also found in Section \ref{subsection:hypo1} with the recency bias. The positive effect of the formality bias, indicated by the positive $\hat{\beta}$s, suggests that, as predicted by our hypothesis, a sentence that deviates from standard written text (excessive punctuation, missing subjects, or run on sentence) increases the likelihood the GPT model will predict it is a feature of AAE. 

The few-shot prompting strategy should not be abandoned just because it suffers from the formality bias, since it was shown to enhance model performance with a complex feature such as Habitual Be (see Section \ref{subsubsection:habbe}). This warrants future work on mitigating this bias by experimenting with the nature and the proportion of the few-shot examples that deviate from standard written text.

\section{Conclusion} %

Large Language Models (LLMs) continue to show promise in automating feature annotation tasks, yet significant challenges remain, particularly when it comes to language varieties with complex grammatical features that are not represented in standard written conventions, such as the Habitual Be in African American English (AAE). 
The main contributions of our work are:

\begin{itemize}[noitemsep,topsep=0pt]
    \item Rule-based and Transformer-based models of AAE grammatical features outperform zero- and few-shot LLMs.
    \item LLMs recognition of AAE is influenced by recency and unrelated features of non-standard written conventions.
    \item We release a Multiple Negation tagger that outperforms LLMs for AAE.
\end{itemize}

We found that, surprisingly, the few-shot approach did not always performed better. It performed worse than the zero-shot approach for Multiple Negation, but better for Habitual Be. We conclude the choice between employing zero-shot and few-shot approaches for classification should be influenced by the complexity of the feature perhaps more than the extent to which relevant data is represented in the LLM's pre-training corpus. 

Our baseline transformer model currently outperforms LLMs in detecting the Habitual Be, but several strategies might enhance LLM performance, especially in light of the hypotheses we have explored. One approach is to mitigate recency bias, a known issue in LLMs, by interleaving or randomizing the order of sentences presented for prediction%
, which can otherwise skew results. Additionally, pre-processing the input data to remove or correct extraneous punctuation and properly segmenting disjointed sentences may help minimize noise and errors when working with something other than mainstream English or transcribed oral speech. 

Our findings provide valuable insights into the potential and limitations of LLMs for AAE. The results underscore the need for further experimentation to assess LLM performance on other linguistic features specific to AAE and transcribed speech. Additionally, exploring how multiple AAE features within a sentence influences the accuracy of LLM predictions could offer a deeper understanding and lead to more robust work on non-standard language varieties in NLP. 

Regarding how these findings could influence the deployment of NLP applications where AAE is spoken, we note that inadequate training of a model to accurately process African American English can result in misrepresentation or difficulty in understanding the speaker's intended message. This may arise from either incoherent transcription outputs or the model's attempt to convert AAE into Mainstream American English (MAE), which can lead to significant errors if the model lacks sufficient training to recognize and handle AAE effectively. Furthermore, an indirect benefit of making AAE-specific annotations more available would be the development of LLMs trained to treat AAE as a distinct language or dialect, thereby contributing to its preservation and recognition as a linguistic standard in its own right.

\section*{Limitations}  

We only gathered transcribed spoken speech data for this experiment. We could have looked at more sources like written texts and speeches from  regions other than Florida. Due to the time and annotated data limitations, we were only able to cover two African American English features for this paper. The  Multiple Negation data with which we have tested the performance of our baseline and LLMs is limited. There is a possibility that the rule-based model could deteriorate in performance with the presence of more data and more variability in the data, requiring more rules to capture different instances of multiple negation. Our regression analysis for recency bias considers the most recent 5 examples. We encoded the feature for this parameter as the proportion of samples in the last 5 predictions that match the current prediction and ran the regression analysis. What this misses is that, for example, where all the past 5 samples match the prediction, and the prediction is correct, there may be a very high coefficient for the recency bias, but we cannot say for certain if it is in fact due to recency bias or due to the LLM correctly predicting the feature. Fortunately, this problem only presented itself when we were experimenting with an extreme ordering of the recency bias examples, such as where all 1s and 0s were grouped together. As such, this is a highly unlikely case. Another limitation we faced was in testing hypothesis 2, due to the augmentation of the habitual be data. The lack of natural sentences structures and meanings made it more difficult to distinguish whether the sentences contained missing subjects or run on aspects. Finally, the last limitation we identify is of the comparison between the LLMs and transformers. The transformers performance may be explained by the fact that they have been trained explicitly for detecting the habitual be feature, while LLMs have not. But then, the goal of this paper was to test the capability of general purpose LLMs to tag AAE features without having to fine-tune the model.

\section*{Ethical considerations}
The involved university does not require IRB approval for this kind of study, which uses publicly available data. 

The materials are used in accordance with SPOHP's guidelines for academic research and educational purposes. Proper credit is given to the program for the original data sources, and the use of these materials complies with the program's policies on non-commercial, educational, and research use. This research uses data from the CORAAL, which is licensed under a \href{https://creativecommons.org/licenses/by-nc-sa/4.0/}{Creative Commons Attribution-NonCommercial 4.0 International License (CC BY-NC 4.0)}. The dataset is used for non-commercial academic research, with proper attribution given to the creators. All usage complies with the terms outlined in the license, ensuring that the data is used for educational and research purposes only. %

The GPT model used in this research were accessed through OpenAI's platform following OpenAI's \href{https://openai.com/policies/terms-of-use/}{terms of service} and \href{https://openai.com/policies/usage-policies/}{usage policies}. The LLaMA models were used under the non-commercial research license provided by Meta, ensuring compliance with the restrictions specified in the \href{https://ai.meta.com/blog/large-language-model-llama-meta-ai/}{LLaMA License Agreement}. The models were used solely for academic research purposes.

We do not see any other concrete risks concerning use of our research results. Of course, in the long run, any research results on AI methods based on large language models could potentially be used in contexts of harmful and unsafe applications of AI. It is possible that feature tagging of a language of specific group of people could be used to identify the demographics of the individuals who use African American English. 

We recognized the ethical and environmental impact of the carbon footprint associated with using LLMs. Our findings show that LLM indeed might not be suitable for solving even a simple tagging task. This result will encourage future researchers to rethink whether LLM can be helpful, considering performance and the chance of generating carbon footprint.

\section*{Acknowledgments}

We thank the Samuel Proctor Oral History Program (SPOHP) at the University of Florida (UF) for their cooperation and for making the data easily accessible to us. We gratefully acknowledge the narrators of the oral histories housed in the Joel Buchanan Archive at UF who used their unique voice to share their own stories with the public.  This research is part of the project ``\href{https://oral.history.ufl.edu/projects/reanimating-african-american-oral-histories-of-the-gulf-south/}{Reanimating African American Histories of the Gulf South}'' which is supported by the National Endowment for the Humanities (PW-277433-21). We thank the anonymous reviewers for their valuable feedback.

\section*{Contribution statement}

SM and KT are the senior and corresponding authors. We follow the CRediT taxonomy\footnote{\url{https://credit.niso.org/}}. Conceptualization: SM, KT; Data curation: RP, AR, JG, PH; Formal Analysis: KT, SM; Funding acquisition: SM, KT; Investigation: SM, KT; Methodology: RP, AR, JG, PH, SM, KT; Project administration: RP; Resources: SM; Software: RP, AR, JG, PH; Supervision: SM, KT; Validation: RP, AR, JG, PH; and Writing – original draft: RP, AR, JG, PH, SM, KT and Writing – review \& editing: AR, PH, SM, KT.

\bibliography{aclanthology,custom}

\appendix

\end{document}